\documentclass[10pt,twocolumn,letterpaper]{article}

\PassOptionsToPackage{disable}{todonotes}
\PassOptionsToPackage{final}{showkeys}

\PassOptionsToPackage{frozencache,cachedir=.}{minted}

\PassOptionsToPackage{round}{natbib}
\usepackage{garyw}

\title{GraB-sampler: Optimal Permutation-based SGD Data Sampler for PyTorch
\thanks{Preprint. Under review. Cornell CS M.Eng. Project. Package publicly available at \url{https://pypi.org/project/grab-sampler/}}
}

\author{
\textbf{Guanghao Wei}
\thanks{Project Advisor: Chris De Sa (\href{mailto:cdesa@cs.cornell.edu}{cdesa@cs.cornell.edu}) of Cornell University}
\\
Department of Computer Science\\
Cornell University\\
{\tt\small gw338@cornell.edu}
}

\DTMlangsetup[en-US]{showdayofmonth=false}
\date{\DTMdate{2023-5-31}}

\begin{document}

\maketitle

\begin{abstract}
The online Gradient Balancing (GraB) algorithm greedily choosing the examples ordering by solving the herding problem using per-sample gradients is proved to be the theoretically optimal solution that guarantees to outperform Random Reshuffling.
However, there is currently no efficient implementation of GraB for the community to easily use it.

This work presents an efficient Python library, \textit{GraB-sampler}, that allows the community to easily use GraB algorithms and proposes 5 variants of the GraB algorithm.
The best performance result of the GraB-sampler reproduces the training loss and test accuracy results while only in the cost of $8.7\%$ training time overhead and $0.85\%$ peak GPU memory usage overhead.
\end{abstract}

\section{Introduction}
\label{sec:intro}

Random Reshuffling(RR), which samples training data examples without replacement, has become the \textit{de facto} example ordering method in modern deep learning libraries.
However, recent research on online Gradient Balancing (GraB)~\citep{Lu2022GraBFP} reveals that there exist permutation-based example orderings that are guaranteed to outperform random reshuffling(RR), and the follow-on work shows that GraB is theoretically optimal \citep{Cha2023TighterLB}.
GraB connects permuted-order SGD to the \textit{herding problem}~\citep{Harvey2014NearOptimalH} that greedily chooses data orderings depending on per-sample gradients to further speed up the convergence of neural network training empirically. 
Empirical study shows that not only does GraB allow fast minimization of the empirical risk, but also lets the model generalize better on multiple deep learning tasks.

The herding problem requires all vectors to be pre-centered to ensure they sum to zero \citep{Lu2022GraBFP}.
The original work of GraB proposes a herding-based online Gradient Balancing algorithm that uses stale gradient means to stimulate the running average of the gradients in the current epoch (Mean Balance).
More recent work in CD-GraB~\citep{cooper2023cdgrab} proposed Pair Balance that further reduces dependencies on stale gradient means and works efficiently under distributed and parallel settings.
In this work, we propose more variants of the balancing subroutine that attempts to solve various issues, namely: Batch Balance, Recursive Balance, and Recursive Pair Balance.

GraB algorithm requires per-sample gradients while solving the herding problem. 
In general, it's hard to implement it in the vanilla PyTorch Automatic Differentiation (AD) framework~\citep{Paszke2017AutomaticDI} because the C++ kernel of PyTorch averages the per-sample gradients within a batch before it is passed back to Python or forwarded to the next layer.
The previous implementation of GraB was based on BackPACK~\citep{Dangel2019BackPACKPM}, a third-party library that builds on the top of PyTorch to compute quantities other than the gradient efficiently.
For other implementations, no efficient solution exists.
One goal of this project is to get rid of third-party library dependencies other than PyTorch and to provide the community with a simple, efficient, and off-the-shelf solution of GraB.

To make it easier for the entire community to use GraB algorithm in their code, my work implements a Python library, \textit{GraB-sampler}, that allows users to use GraB with a minimum of 3 line changes to their training script.

As a CS M.Eng. Project, my work includes:
\begin{enumerate}[nosep]
    \item Implement the PyTorch data loader compatible sampler that supports 5 balance algorithms, namely:
    \begin{enumerate}[nosep]
        \item Mean Balance (Vanilla GraB~\citep{Lu2022GraBFP})
        \item Pair Balance (CD-GraB~\citep{cooper2023cdgrab})
        \item Batch Balance
        \item Recursive Balance
        \item Recursive Pair Balance
    \end{enumerate}
    equipped with 2 alternative Balancing kernels:
    \begin{enumerate}[nosep]
        \item Deterministic Balancing
        \item Probabilistic Balancing with Logarithm Bound \citep{Lu2022GraBFP,Alweiss2020DiscrepancyMV}
    \end{enumerate}
    and other functional requirements that will be discussed in \cref{sec:method}.
    \item Reproduce the LeNet on CIFAR-10 experiments and performance of the original paper.
    \item Benchmark the performance of all balance algorithms.
\end{enumerate}

The library is now released on \href{https://test.pypi.org/project/grab-sampler/}{PyPI}.

\section{Preliminaries and GraB Variants}
\label{sec:related}

While the vanilla GraB algorithm refers to the herding-based online Gradient Balancing algorithm using the stale mean of sample gradients \citep{Lu2022GraBFP} and Pair Balance refers to the parallel-variant of it that works efficiently under the distributed and parallel setting \citep{cooper2023cdgrab}, we also proposed Batch Balance, Recursive Balance, and Recursive Pair Balance that designed under different desires.
Considering the purpose of this report, this section gives a preliminary explanation of these variants at a high level, and more prescriptive details will be included in the future workshop paper.

\paragraph{Mean Balance (Vanilla GraB)} Mean Balance is the vanilla GraB algorithm that uses the stale gradient means to stimulate the running average of gradients and then solves the herding problem by assigning all examples with either \textbf{+} or \textbf{-} sign while computing the balancing.
Mean Balance takes $\bigO{n}$ computation and $\bigO{d}$ memory overhead for storing the stale mean and accumulating vector.

\paragraph{Pair Balance (CD-GraB)} Instead of using a pre-centered vector, the centralized version of Pair Balance uses the difference between 2 vectors to compute the balancing and assigns \textbf{+} to one example and assigns \textbf{-} to the other one.

\paragraph{Batch Balance} Batch Balance is designed the seek more parallelism while computing the balancing across the batch.
Batch balance delays updating the accumulator vector until the balancing is calculated for a full batch, which makes all example gradients within the same batch relatively independent of each other while computing the balancing, bringing the potential of parallelism.

\paragraph{Recursive Balance} One issue of using GraB in practice is that GraB requires various epochs (usually $\ge 10$) of training and reordering to witness the performance gain compared with RR.
However, it is almost unaffordable to train or fine-tune an LLM with billions of parameters for many epochs in practice.
Recursive Balance, inspired by \citet{Dwivedi2021KernelT}, taking advantage of a binary-tree structure of accumulator-balancing computation, balances each example $D$ times, where $D$ is the depth of the recursive tree, within a single epoch.

Empirically, Recursive Balance results in even faster convergence, especially in the first few epochs.
However, the memory overhead is now exponential to $D$ as of $\bigO{2^Dd}$, which becomes a bottleneck of applying GraB to tremendous models.
The computation overhead also scales as $\bigO{Dn}$ over RR.

\paragraph{Recursive Pair Balance} Recursive Pair Balance is designed to seek all the goods from all variants.
Recursive Pair Balance uses the difference between 2 example gradients to compute the balancing sign, so there is less memory overhead without the need to store the stale mean vector.
It also uses a tree structure of accumulators to achieve faster convergence.
Last but not least, it assumes that the accumulator won't be updated within a batch, which enables the possibility of implementing highly parallelism codes to further leverage performance.

However, Recursive Pair Balance requires the batch size to be a perfect power of 2, which is not a big issue in practice because people used to choose small batch sizes like 16, 64, or large batch size 1024, which are all power of 2.
But the $\bigO{2^Dd}$ memory overhead is inevitable.

\section{GraB-sampler}
\label{sec:method}

GraB-sampler is the Python package of an efficient PyTorch-based sampler that supports all 5 GraB-style example ordering algorithms mentioned above.
This section talks about some design choices and features of the implementation.

\subsection{Native Support PyTorch}

GraB-sampler inherited \textit{torch.utils.data.Sampler}, so it natively supports PyTorch DataSet and DataLoader.
Since the data permutation depends on the per-sample gradients in the GraB algorithm, passing these gradients to the sampler during training is necessary.

A minimum code snippet that uses our library by only changing 3 lines of code shows as the following.

\begin{minted}
[
frame=lines,
framesep=2mm,
baselinestretch=1.2,
bgcolor=lightgray,
fontsize=\footnotesize,
]
{python}
# Initiate model, params, dataset
...
sampler = GraBSampler(dataset, params)
dataloader = torch.utils.data.DataLoader(
    dataset, sampler=sampler
)
...
# Train loop begin
for epoch in range(epochs):
    for x, y in dataloader:
        # Get per-sample gradients and loss
        ...
        sampler.step(ft_per_sample_grads)
        ...
        # Update optimizer for backpropogation
        ...
\end{minted}

The core component of the GraB-sampler is a \textit{sorter}. 
The \textit{sorter} updates the GraB algorithm based on the per-sampler gradients passed in and generates a new data permutation at the beginning of each epoch.

\subsection{Functional Programming}

All variants of the GraB algorithms require computing per-sample gradients to compute the balancing signs. 
In general, it's hard to implement it in the vanilla PyTorch Automatic Differentiation (AD) framework \citep{Paszke2017AutomaticDI} because the C++ kernel of PyTorch averages the per-sample gradients within a batch before it is passed back to Python or forwarded to the next layer.

Fortunately, the recently released PyTorch 2.0 integrates Functorch which supports the efficient computation of Per-sample Gradients. 
Alas, it requires a functional programming style of coding and requires the model to be pure functional procedures, disallowing Neural Network layers including randomness (Dropout) or storing inter-batch statistics (BatchNorm).

\begin{figure*}[htbp]
\begin{center}
\includegraphics[width=\linewidth]{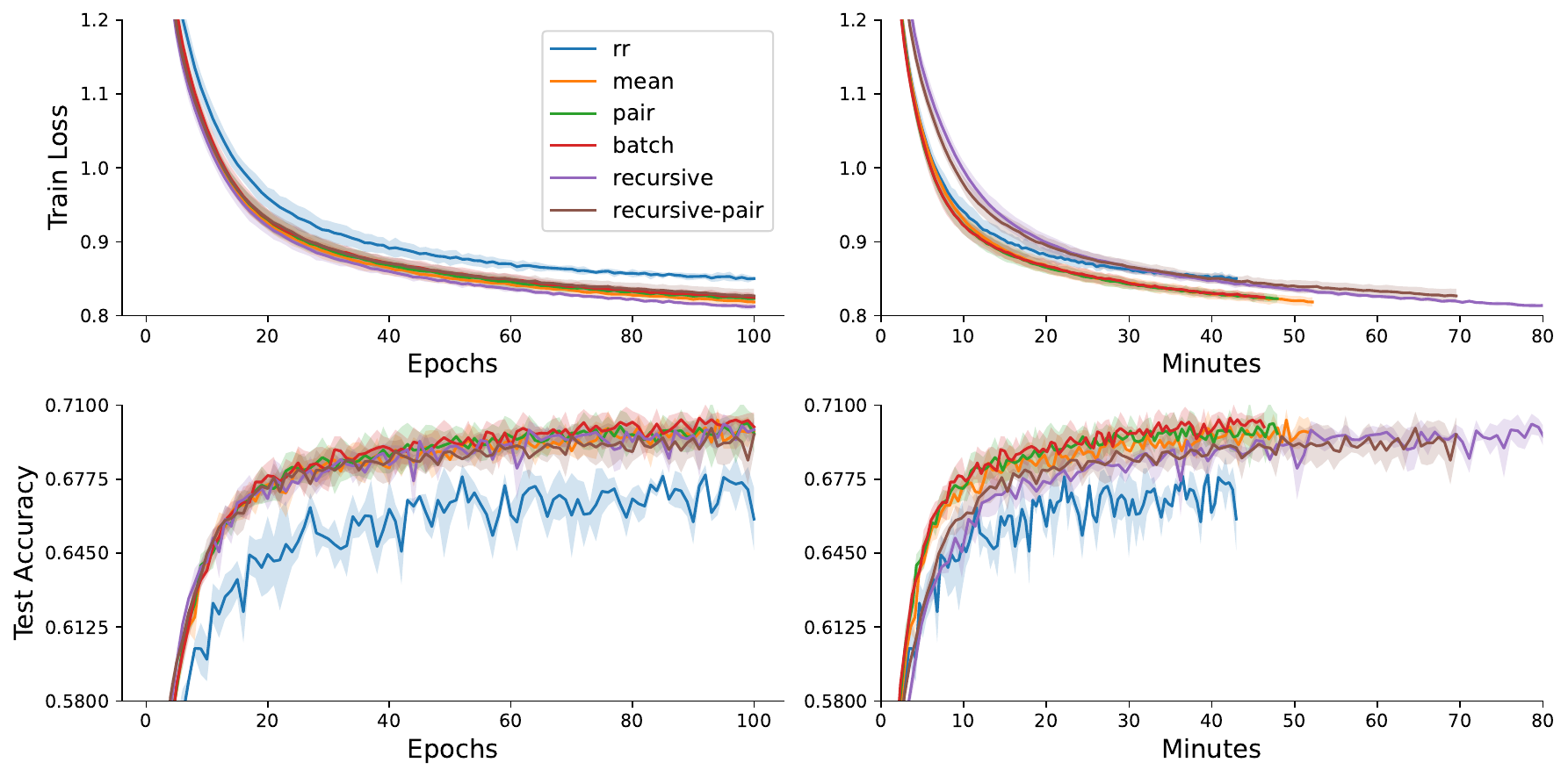}
\end{center}
\caption{
\textbf{Reproduce Experiments: LeNet on CIFAR10 -- }
All experiments are training for 100 epochs, repeated 3 times under seeds 0, 7, and 42. For Recursive Balance and Recursive Pair Balance, depth $D=5$.
Upon the first 5 epochs, both recursive-base balances outperform all the other methods.
All 5 variants converge at a similar performance after 100 epochs, but both recursive-base balances have remarkable overhead in terms of training time.
}
\label{fig:cifar10}
\end{figure*}

\begin{table*}[htbp]
\centering
\resizebox{\linewidth}{!}{ %
\begin{tabular}{c|cccccc}
                                                                                        & \textbf{Train Loss} & \textbf{\begin{tabular}[c]{@{}c@{}}Train Loss\\ (after 5 epochs)\end{tabular}} & \textbf{Test Accuracy} & \textbf{\begin{tabular}[c]{@{}c@{}}Test Accuracy\\ (after 5 epochs)\end{tabular}} & \textbf{\begin{tabular}[c]{@{}c@{}}Training Time\\ (seconds per epoch)\end{tabular}} & \textbf{\begin{tabular}[c]{@{}c@{}}Peak GPU Memory\\ Allocation (MB)\end{tabular}} \\ \hline
\textbf{\begin{tabular}[c]{@{}c@{}}Random Reshuffling\\ (Classic PyTorch)\end{tabular}} & 0.8459              & 1.218                                                                          & 0.6615                 & 0.5765                                                                            & 16.49                                                                                & 18.613                                                                             \\
\textbf{\begin{tabular}[c]{@{}c@{}}Random Reshuffling\\ (Functorch)\end{tabular}}       & 0.8503              & 1.2436                                                                         & 0.66                   & 0.5627                                                                            & 25.77                                                                                & 26.9                                                                               \\ \hline
\textbf{Mean Balance}                                                                   & 0.8186              & 1.2019                                                                         & 0.6986                 & 0.5848                                                                            & 31.29                                                                                & 27.6                                                                               \\
\textbf{Pair Balance}                                                                   & 0.8232              & 1.2096                                                                         & 0.6976                 & 0.5850                                                                            & 28.75                                                                                & \textbf{27.1}                                                                      \\
\textbf{Batch Balance}                                                                  & 0.8248              & 1.2122                                                                         & \textbf{0.7005}        & 0.5870                                                                            & \textbf{27.78}                                                                       & 30.3                                                                               \\
\textbf{Recursive Balance}                                                              & \textbf{0.8125}     & \textbf{1.1814}                                                                & 0.6993                 & 0.5956                                                                            & 49.69                                                                                & 44.7                                                                               \\
\textbf{Recursive Pair Balance}                                                         & 0.8268              & 1.1909                                                                         & 0.6971                 & \textbf{0.5968}                                                                   & 41.72                                                                                & 41.3                                                                              
\end{tabular}
} %
\caption{
\textbf{Reproduce Experiments: LeNet on CIFAR10 -- }
All experiments are training for 100 epochs, repeated 3 times under seeds 0, 7, and 42. For Recursive Balance and Recursive Pair Balance, depth $D=5$.
Only the sample means are reported.
The classic PyTorch run follows the tutorial on the PyTorch website to simulate a classical use of PyTorch.
Regardless of randomness reproducibility, 2 RR experiments are supposed to be equivalent to each other.
} %
\label{tab:cifar10_tab}
\end{table*}

\begin{figure*}[htbp]
\begin{center}
\includegraphics[width=\linewidth]{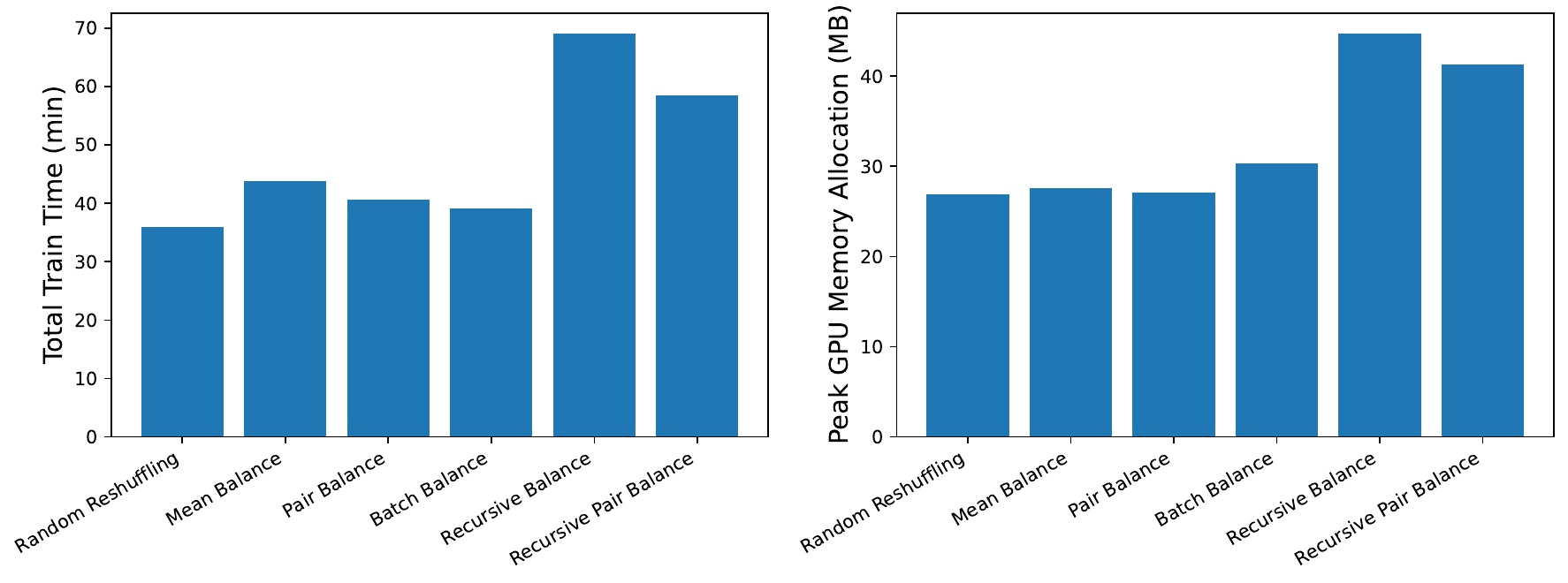}
\end{center}
\caption{
\textbf{Train Time and GPU Memory Usage Benchmark -- }
For Recursive Balance and Recursive Pair Balance, depth $D=5$.
Mean Balance, Pair Balance, and Batch Balance only result in $8.7\% \sim 22\%$ overhead in terms of training time and $0.85\% \sim 12\%$ overhead in terms of peak GPU memory allocation.
However, the recursive-based balancing results in $60\% \sim 90\%$ in terms of training time and $53\% \sim 66\%$ overhead in terms of peak GPU memory allocation.
}
\label{fig:time_mem}
\end{figure*}

\section{Evaluation}
\label{sec:results}

In this section, to benchmark the performance and check the correctness of the implementation, I reproduced the CIFAR10~\citep{Krizhevsky2012ImageNetCW} experiments of all GraB~\citep{Lu2022GraBFP} variants by training a LeNet~\citep{LeCun1998GradientbasedLA}.
All the experiments run on an instance configured with a 12-core AMD Ryzen 1920X 3.5GHz CPU, 64GB memory, and an NVIDIA GeForce RTX 2060 GPU.
All the hyper-parameters are the same as the original paper, namely
\begin{itemize}[nosep]
    \item SGD optimizer
    \item Batch size: 16
    \item Learning rate: 0.001
    \item Weight decay: 0.01
    \item Momentum: 0.9
\end{itemize}

In order to reduce the bias caused by a particular seed, each experiment is run 3 times with seeds 0, 7, and 42.

\paragraph{Model and Dataset} The CIFAR10 dataset consists of 60,000 32x32 color images in 10 classes, with 6,000 images per class.
There are 50,000 training images and 10,000 test images.
LeNet is a classic convolutional neural network proposed by \citet{LeCun1998GradientbasedLA}.
LeNet contains 62,006 parameters.

\paragraph{Reproduce Experiments} \cref{tab:cifar10_tab} and \cref{fig:cifar10} shows the results of reproducing the LeNet on CIFAR10 experiments with all variants of GraB-samplers.
\cref{fig:time_mem} compares the overhead of each variant with the RR sampler in terms of training time and peak GPU memory allocation.

The experiments greatly reproduce the result of \citet{Lu2022GraBFP} that GraB outperforms RR.
Additionally, Recursive Balance and Recursive Pair Balance shows better convergence rate in the first 10 epochs both in Training loss and Test accuracy.
However, both recursive-base balances have remarkable overhead in terms of training time and GPU memory usage.

\section{Conclusions}

In this work, I present a Python library, \textit{GraB-sampler}, that allows users to easily use 5 GraB variants balance algorithms and 2 alternative balancing kernels.
Among the 5 variants, Batch Balance, Recursive Balance, and Recursive Pair Balance are newly proposed.
I reproduce the LeNet on CIFAR10 experiments to benchmark the performance and check the correctness of the implementation.

\section{Acknowledgement}

This work and the research behind it are conducted within the Cornell Relax ML Lab lead by Prof. Chris De Sa.
This is the following work of Yucheng Lu and Wentao Guo's previous work on GraB~\citep{Lu2022GraBFP} and CD-GraB~\citep{cooper2023cdgrab}.
I am grateful for working closely with Wentao Guo on this project.

{
    \clearpage
    \small
    \renewcommand{\bibname}{References}
    \bibliographystyle{abbrvnat}
    \bibliography{main}
}

\end{document}